\documentclass[letterpaper, 10 pt, conference]{ieeeconf}  

\IEEEoverridecommandlockouts                              

\usepackage{graphics} 
\usepackage{epsfig} 
\usepackage{float} 

\usepackage{mathptmx} 
\usepackage{times} 
\usepackage{amsmath} 
\usepackage{amssymb}  

\usepackage{dsfont}
\usepackage{soul}
\usepackage[normalem]{ulem}
\usepackage[utf8]{inputenc}
\usepackage[small]{caption}
\usepackage{graphicx}

\usepackage{amsmath}
\usepackage{booktabs}
\usepackage{algorithm}
\usepackage[noend]{algorithmic}
\usepackage{amsfonts}
\usepackage{blindtext}
\usepackage{subfig}
\graphicspath{{Figures//}}
\DeclareGraphicsExtensions{.eps,.pdf,.png,.jpg,.tif,.tiff,.ps}

\usepackage{caption}
\usepackage{tabu}

\usepackage{outlines}

\usepackage{breqn}

\makeatletter
\let\NAT@parse\undefined
\makeatother
\usepackage[hidelinks]{hyperref}  
\usepackage{url}
\usepackage[nameinlink,capitalize]{cleveref}
\usepackage{soul}
\usepackage[colorinlistoftodos]{todonotes}

\definecolor{navy}{rgb}{0.1, 0.1, 0.8}
\definecolor{gray}{rgb}{0.6, 0.6, 0.6}
\definecolor{myblue}{rgb}{.8, .8, 1}
\definecolor{olive}{rgb}{0.1, 0.5, 0.1}
\definecolor{magenta}{rgb}{0.55, 0.0, 0.55}
\usepackage{adjustbox}
\usepackage{textcomp}
\usepackage{siunitx}
\usepackage{subfig}

\usepackage{lipsum}

\usepackage{xspace}

\newcommand{\citet}[1]{\citeauthor{#1}~\shortcite{#1}}
\newcommand{\citep}{\cite}

\newcommand{\titlename}{Traffic incident duration prediction via a deep learning framework for text description encoding}

\title{\LARGE \bf \titlename}

\author{Artur Grigorev $^{1,*}$, Adriana-Simona~Mih\u{a}i\c{t}\u{a} $^{1}$, Khaled Saleh$^{1}$, Massimo Piccardi $^{1}$
\thanks{$^{1}$ University of Technology, Sydney}
\thanks{$^{*}$ Corresponding author: Artur.Grigorev@student.uts.edu.au (A. Grigorev)} 
}

\begin{document}

\maketitle
\thispagestyle{empty}
\pagestyle{empty}

\begin{abstract}
Predicting the traffic incident duration is a hard problem to solve due to the stochastic nature of incident occurrence in space and time, a lack of information at the beginning of a reported traffic disruption, and lack of advanced methods in transport engineering to derive insights from past accidents.
This paper proposes a new fusion framework for predicting the incident duration from limited information by using an integration of machine learning with traffic flow/speed and incident description as features, encoded via several Deep Learning methods (ANN autoencoder and character-level LSTM-ANN sentiment classifier). The paper constructs a cross-disciplinary modelling approach in transport and data science. The approach improves the incident duration prediction accuracy over the top-performing ML models applied to baseline incident reports. Results show that our proposed method can improve the accuracy by $60\%$ when compared to standard linear or support vector regression models, and a further $7\%$ improvement with respect to the hybrid deep learning auto-encoded GBDT model which seems to outperform all other models. The application area is the city of San Francisco, rich in both traffic incident logs (Countrywide Traffic Accident Data set) and past historical traffic congestion information (5-minute precision measurements from Caltrans Performance Measurement System).
\end{abstract}



\section{INTRODUCTION}\label{introduction}

When traffic accidents occur, the majority of traffic management centres (TMCs) store a brief textual description and the GPS coordinates of the incident. There is a lot of uncertainty at the beginning of disruptions with regards to how long the traffic incident will last, and most of the time, centres do not have an overview of the length or severity of the disruptions. Therefore, it is extremely insightful for TMCs to be able to utilise the data on historical traffic flows or readily available accident description to predict or improve predictions of the incident duration. In order to improve predictions, we need more information on the factors (both readily-available and historical) which can have an effect on the incident duration prediction accuracy. This paper presents an advanced incident duration prediction framework which makes use of additional incident report variables and past incidents records, merged into a hybrid machine learning (ML) modelling approach with deep learning encoding of additional features (e.g. textual incident description and historical traffic flow in the vicinity of the section). Feature encoding is justified since the traffic incident description and traffic flow/speed measurements have a high dimensionality, which can lead to overfitting when using ML models and it may be worsened by the small size of a typical incident report data set.

This paper is organised as follows: Section \ref{introduction} presents the challenges and reviews the related works; Section \ref{data_sources} introduces the data sources we have used as well as our traffic flow mapping algorithm for feature construction; Section \ref{methodology} proposes our modelling framework and explains the ML models we have used, the LSTM sentiment encoder for textual incident descriptions, and the ANN encoder for traffic flow speed; Section \ref{results} introduces the results before summarising all findings in the Conclusions section.

\newcommand{\selfcite}{$[$\textit{self-citation invisible due to double-blind review}$]$}

\subsection{RELATED WORK}\label{realted}

There are multiple research papers which use baseline incident reports from TMC with different machine learning models to predict the traffic incident duration \cite{li2018overview}. The use of traffic flow and incident description features is found to be rare and mostly specific - topical text modelling \cite{twitter1} for the task of the incident duration detection, modelling or incident impact prediction by using traffic flows \cite{fukuda2020short}. And its scarcity is highlighted since it requires the involvement of additional specific models with a feature fusion approach. In other words, traffic flow data is rarely combined with textual incident description and an actual incident reports since it requires a higher system complexity.

But feature combination can be observed in some specific research studies related to the traffic incident impact prediction, which rely heavily on the historical traffic flow data with and without consideration of features that are describing the incident \cite{fukuda2020short}; other works have addressed a similar approach \cite{WenTRB2018,Mihaita2019,Mihaita2020}. Also, these works don't focus on the incident duration prediction.

Sometimes, researchers try to apply uniform ML approaches or specific models for all the sub-tasks. Separate RBM models were applied  to  different  kinds  of  features  and  feature  fusion representing a uniform application of ML method to different data sets \cite{li2020deep}. Also, kNN and Bayesian cost-sensitive networks were combined for the task of the incident duration prediction \cite{kuang2019predicting}. But neither of these research studies investigated a deep dive into their model selection.

Since we have the incident description and incident severity values in our incident reports, we can utilise specific models for the task of sentiment classification. Previously, the LSTM architecture has been compared with Support Vector Machines, Artificial Neural Networks, Deep Belief Networks and Latent Dirichlet Association on the task of detection of incidents from social media data \cite{zhang2018deep}. LSTM was also successfully used for stock price prediction \cite{sen2018stock}, making it applicable for modelling of traffic flow/speed time-series data. Despite its superior performance, we need to uplift and bring significant modifications to this architecture. Since we are planning to use encoded time series with machine learning methods, we need a controllable size of the feature vector to simultaneously avoid overfitting and provide enough information for ML methods. This is why we propose to use LSTM coupled with ANN, where the ANN feature vector size and the activation function are varied.

\section{CASE STUDY}\label{data_sources}

In this study we assume that textual incident reports as well as historical traffic flows and speed data (including the ones from the moment when an incident happened) are readily available at the moment the incident was reported and sufficient to make the prediction of its duration. 

\subsection{Incident description data set and baseline feature set}

A Countrywide Traffic Accident Data set (CTADS) has been recently published \cite{moosavi2019accident}-\cite{moosavi2019countrywide}, which contains about 1.5 million traffic accident records across 49 states of United States of America from February 2016 to December 2020 (version 4). Each incident report contains 47 features describing the traffic accident. The majority of these traffic accidents were recorded in the state of California. The most notable features include: a) Incident Severity (valued from 1 to 4), b) Start and End Time of the incident (from which the traffic incident duration is derivable), c) The road extent affected by the accident, d) textual Incident Description, d) weather and lighting conditions. For the extended description of features please refer to the original paper describing the data set \cite{moosavi2019countrywide}. This data set allows us to use the textual incident description and, hence, apply a sentiment analysis methodology (based on  the incident severity) \cite{alkheder2017severity}. We further refer to these features as a baseline feature set, excluding the textual incident description.

\subsection{Traffic flow and speed data}

\begin{figure}[!ht]
\centering
\includegraphics[width=0.37\textwidth]{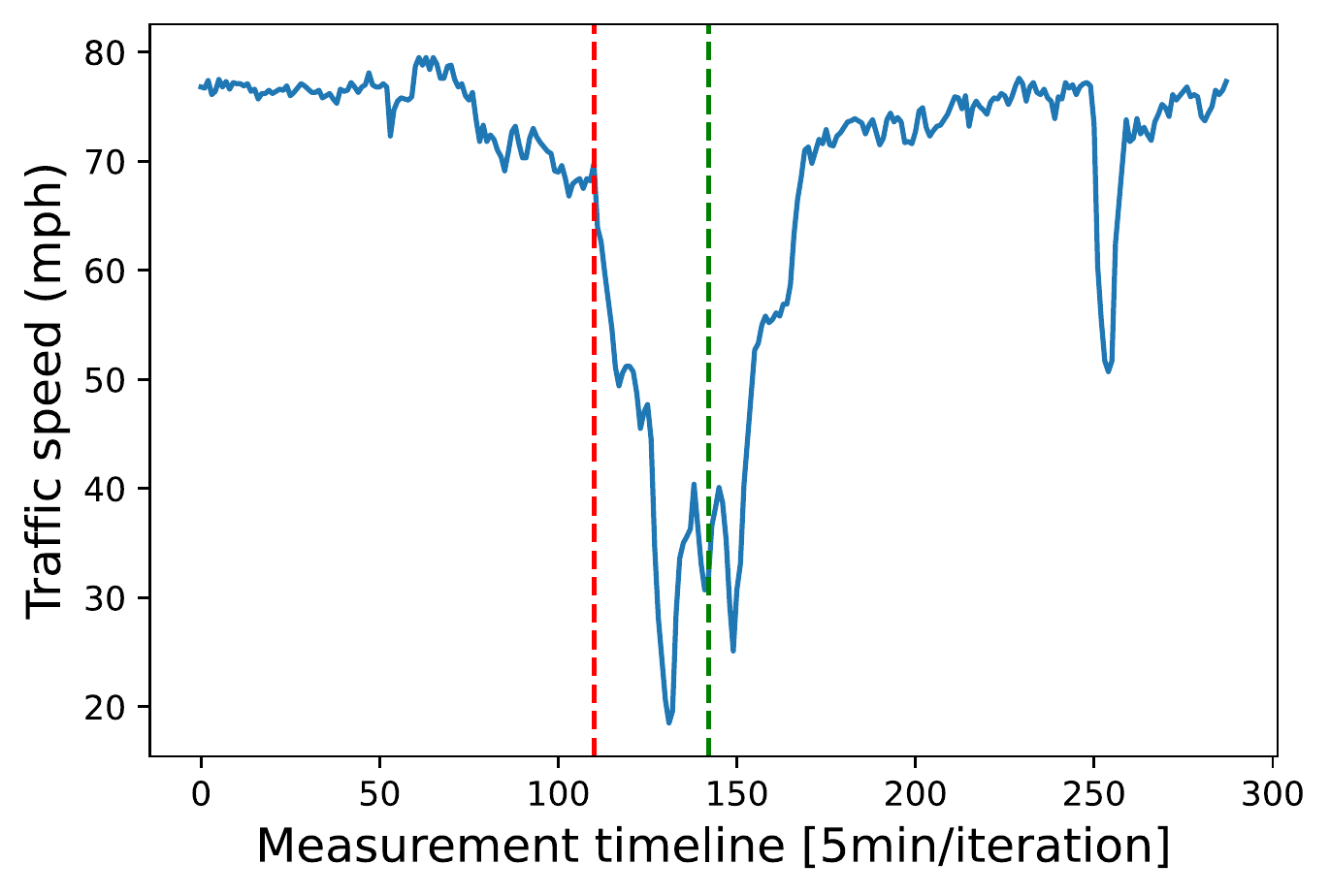}

\includegraphics[width=0.37\textwidth]{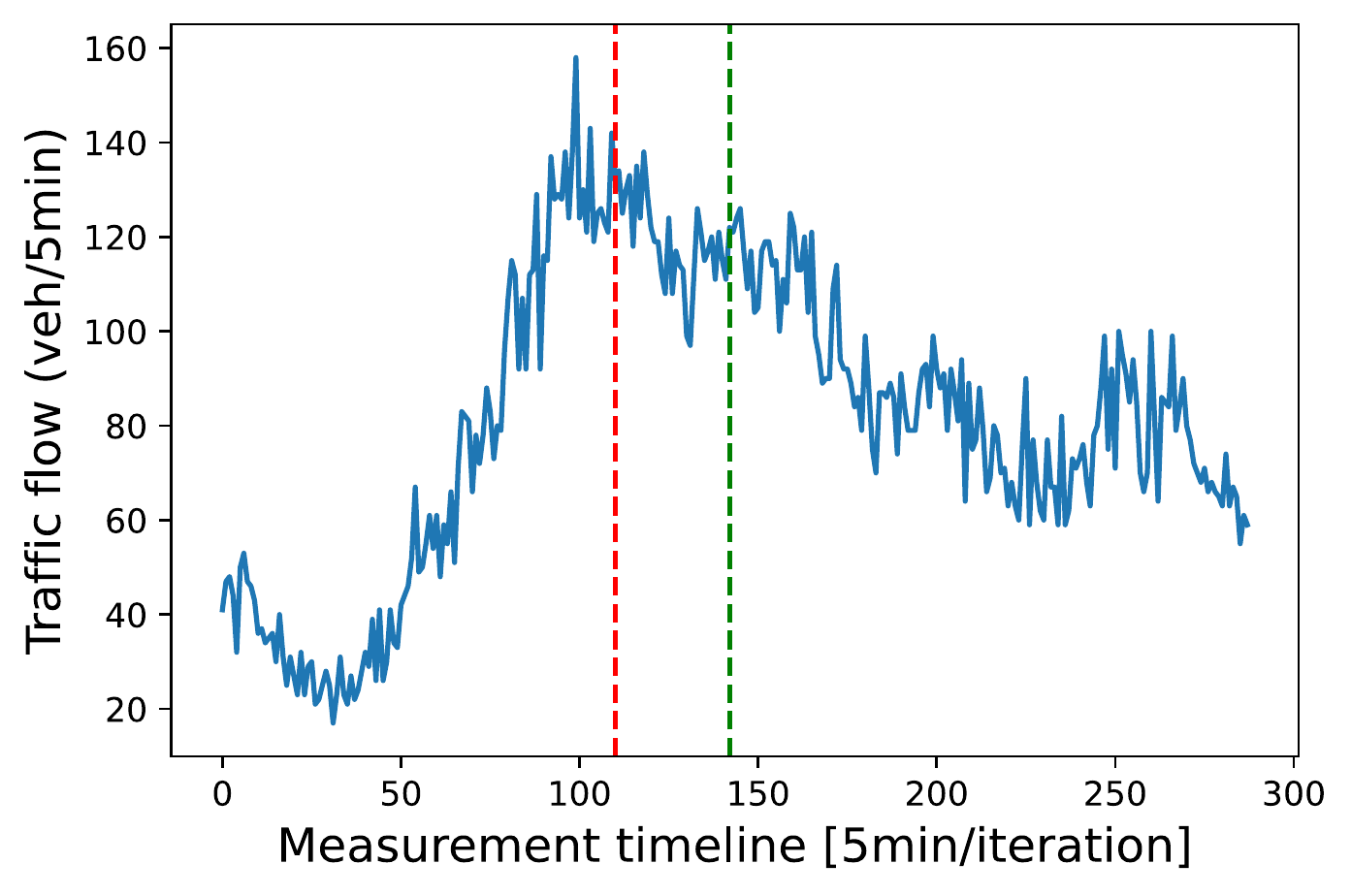}
\caption{ a) Traffic speed and b) Traffic flow plots for the VDS associated to incident A-4798 (accident on US-101 Southbound with duration of 31 5-minute iterations - actual reported incident clearance time, without considering the incident recovery time). The red line denotes the start of the accident, and the green line the end of the accident. The blue line denotes the speed evolution in the vicinity of the incident location (drops almost to 20km/h) while the flow is still running at high values due to large numbers of vehicles blocked in traffic.}
\label{fig:vds}
\vspace{-0.5cm}
\end{figure}

\begin{figure*}[h]
\centering
\includegraphics[width=0.75\textwidth]{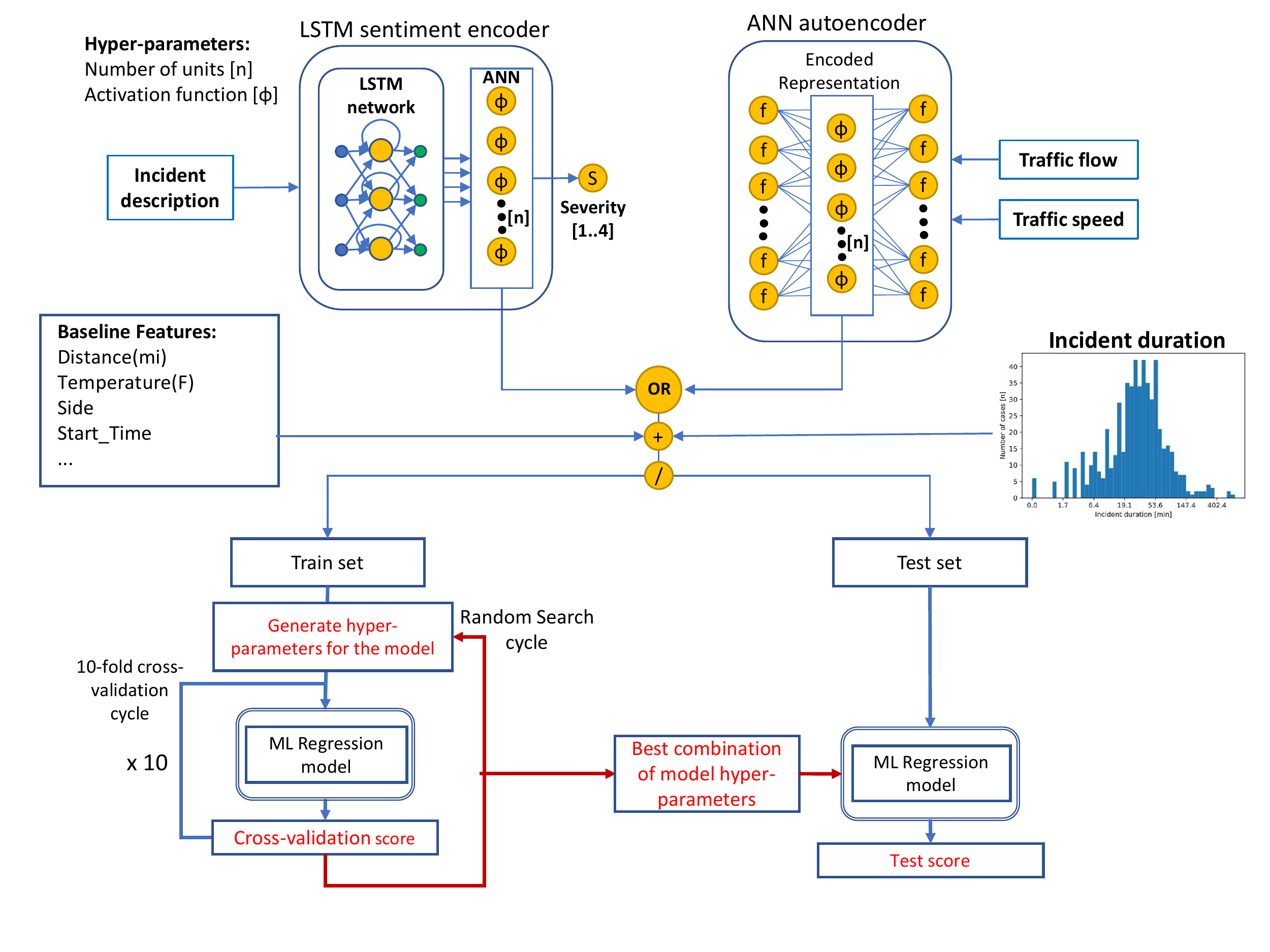}
\centering
\caption{The structure of the proposed framework}
\label{Fig_all}
\end{figure*}

To collect the data on traffic flows and speed we rely on the Caltrans Performance Measurement System (PeMS) \cite{chen2001freeway}, which provides aggregated 5-minute precision measurements of traffic movements across California. Although there is a lot more data for the Los Angeles area (which may be considered in our future research), we decided to concentrate on the area of the city of San Francisco. We focus on 83 Vehicle Detection Stations (VDS) placed in that area, and we try to manually associate each incident occurred in that area with a VDS in their 500m proximity. VDS in PeMS may have detector failures and incomplete readings, which is common across the data set and should be taken into account. Even though the PeMS data set contains data on reported incidents, we decided to use the descriptions from the Countrywise Traffic Accident Data Set since it provides a high-quality description of each incident (47 features in each incident report) extracted from Bing and MapQuest services. 

In total, from 9,275 incidents in the area (extracted from CTADS) we have obtained 1,932 traffic incident reports in a 500m proximity next to VDS stations, which we were able to associate with the correct (no detector faults) and complete traffic flows and speed readings. Incident to VDS association is necessary since both are represented as points and it is not clear which incident is related to which detector since incidents on different separate roads can be in proximity of one detector (also, since we have a different representation of street names in VDS and the incident data sets). The task of VDS-to-incident assignment can be a topic for additional research, but in this paper we summarize our extracted mapping strategy as follows. We extract the following speed and flow readings from each VDS station:

1. Speed – Traffic Speed from the 24h leading to the incident occurrence.

2. Flow – Traffic Flow from the 24h leading to the incident occurrence.

3. Speed7 – Traffic Speed on the same weekday, the week before the incident.

4. Flow7 – Traffic Flow on the same weekday, the week before the incident.

5. SD – the vector difference between the traffic speed on the day of the incident and on the same weekday, the week before the incident.

6. FD – the vector difference between the traffic flow on the day of the incident and on the same weekday, the week before the incident.

Each of these feature vectors contains 288 values, which correspond to 5-minute readings throughout the day. Since each of these vectors have a high dimensionality, we decide to perform dimensionality reduction via an ANN autoencoder.

The use of dimensionality reduction is justified since a large number of explanatory variables can cause model overfitting \cite{sawalha2006transferability,sahraei2021}.

\textbf{Regarding the 288 input values on the day of incident:} the traffic data is taken from the time between the incident start and minus 24h before of its occurrence and not during the entire day after the incident has been lodged.

Figure \ref{fig:vds} shows an example of a traffic speed drop during the incident A-4798. After we have analysed different traffic flows and speed plots we expect that the traffic speed will be the most useful single feature for the task of incident duration prediction as the traffic flow measurement seems to be not affected by the accident (as the majority of vehicles will be waiting for the congestion to clear off the road, and will still be counted as part of the traffic flow). We will also use speed measurements from the weekday, 7 days before the incident in order to obtain the complete picture between what is a regular traffic flow condition versus disrupted traffic condition on the same time and same day of the week. We make the observation that we have also conducted a detailed feature ranking and selection (via SHAP values, forward feature selection, etc.) to several incident data sets which are not presented here due to space limitations.

The point A-4798 point was selected just as an example for a traffic speed drop and its usefulness to the prediction problem; in reality, we have analysed about 100 traffic flow and speed plots before drawing the conclusions (we provide several shapshots of flow and speed reading in the supplementary material~\cite{appendix}). As an observation, by adding severity classification probabilities (from the LSTM-ANN model) to the feature vector for the task of incident duration prediction doesn't seem to be useful since we already included Severity, which is a strong feature.

\section{Methodology}\label{methodology}

Figure \ref{Fig_all} shows how we use the data to perform the incident duration prediction. We combine the baseline feature set with either the encoded textual description or the encoded traffic flow/speed values. The encoder parts of both LSTM-ANN network and the ANN autoencoder have hyper-parameters in the form of number of units and used activation functions to ensure an optimal encoding for the specific ML method. 
After obtaining encoded representations associated with the incident, we search for the optimal hyper-parameters for each ML regression model at each case of the encoded representation. It allows us to adapt the model parameters to work with encoded data and provide the best cross-validation results.

\subsection{LSTM-ANN for the textual incident description encoding}\label{LSTM_ae}

Textual Incident Description in the CTADS data set describes type of disruption caused by the incident and/or location (\cref{table-id}).

\begin{table*}[t]
\small
\begin{tabular*}{\textwidth}{|l|}

Accident on I-280 Northbound at Exit 57 King St.\\
Right hand shoulder blocked due to accident on I-280 Northbound after Exits 54 54A 54B US-101.\\
Lane blocked due to accident on US-101 Presidio Pkwy Southbound at Exit 438 CA-1.\\
Accident on I-80 Westbound at Exits 1 1C / Bryant St / 8th St.\\
Second lane blocked due to accident on I-80 Eastbound at Exits 2B 2C Harrison St.\\
Lane blocked due to accident on US-101 Golden Gate Brg Southbound at Exit 439 Transit Transfer Facility.\\
Right hand shoulder blocked due to accident on I-280 Northbound at Exit 52 San Jose Ave.\\
Right hand shoulder blocked due to accident on US-101 Southbound at Exits 429B 429C Bay Shore Blvd.\\
Lane blocked on exit ramp due to accident on I-280 Northbound at Exit 55 Cesar Chavez.\\
Right hand shoulder blocked due to accident on I-280 Northbound at Ocean Ave.

\end{tabular*}
\caption{Example of the Incident Description values}
\label{table-id}
\end{table*}

\begin{figure}[!h]
\centering
\includegraphics[width=0.40\textwidth]{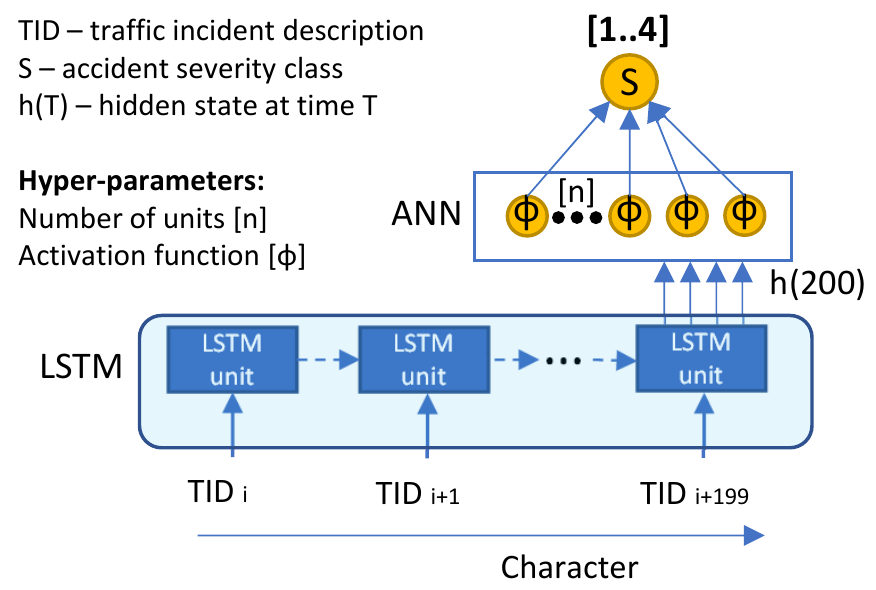}
\centering
\caption{LSTM sentiment encoder structure.}
\label{Fig_lstmann}
\vspace{-0.5cm}
\end{figure}

To perform the encoding of the textual description of the incident we use a combination of character-level LSTM and ANN for the sentiment analysis (Figure \ref{Fig_lstmann}). We use the textual incident description from all the available traffic incident reports for the San Francisco area (9,275 incident records). 
Firstly, we set the target variable for the LSTM classification model as the incident severity (values 1 to 4).

Secondly, we use the encoded representation of the textual description extracted from the LSTM sentiment classification model to use it as additional features for the task of incident duration prediction.

The incident description text is only provided at the beginning of the incident reporting timeline, and no temporal evolution is found across multiple countries for which we analysed the incident logs in our previous work \cite{ag2022}.

Each textual description is formed into repeated strings up to 200 characters in length and each character in that string is then encoded by using binary encoding. 

\begin{figure}
\centering
\includegraphics[width=0.45\textwidth]{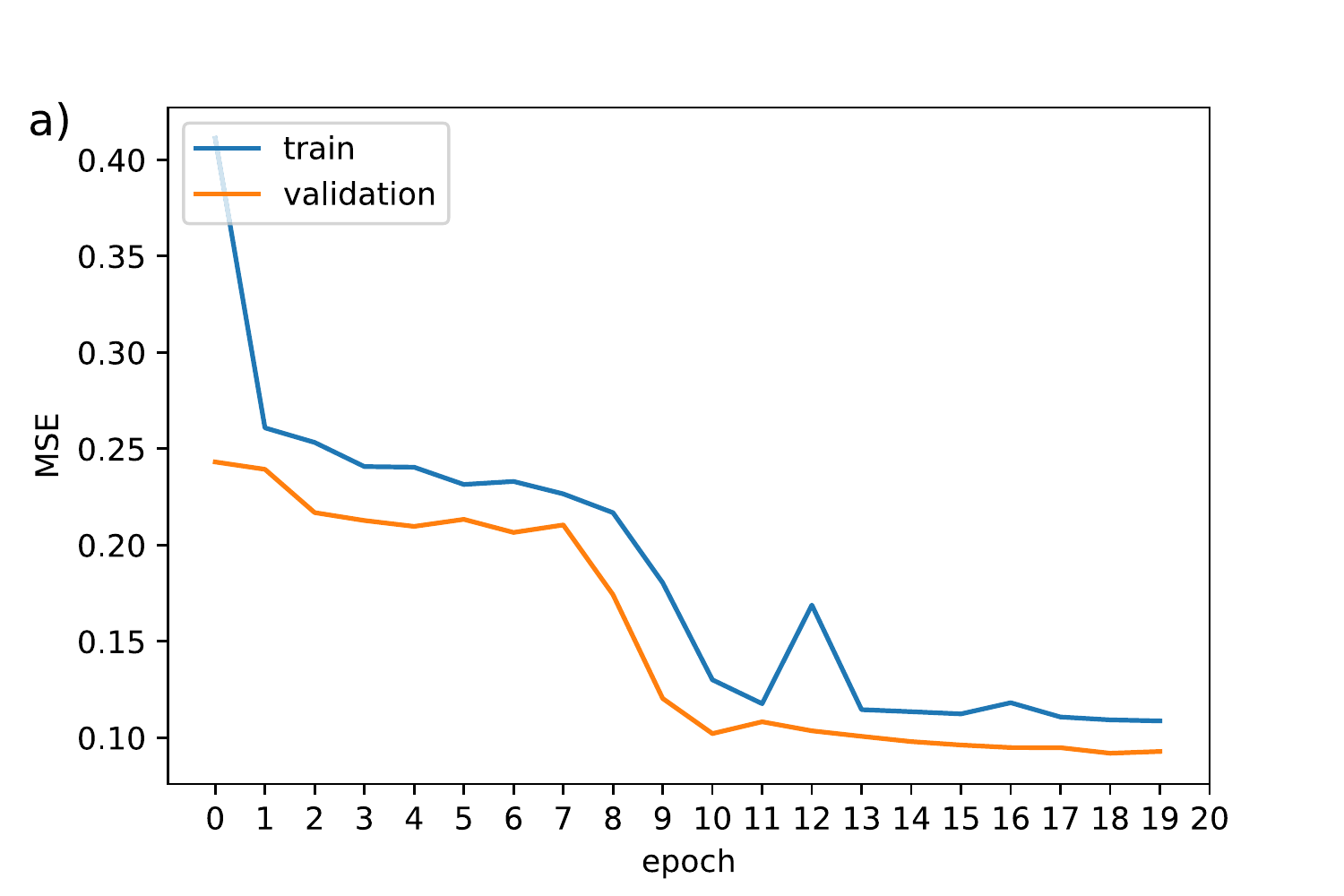}
\caption{Example of LSTM network training results using 12 units, a ReLU activation function, 10 epochs, 80 hidden units. a) Train-validation score over 20 epochs}
\label{Fig_example_1}
\end{figure}

In order to showcase the importance of the textual incident description for the tasks of incident duration prediction and incident severity classification, we perform a word importance analysis using the LIME method (provided in the supplementary material ~\cite{appendix}). We further train a an LSTM model with 80-units hidden state vector. We use the encoding of the incident description by using different numbers of neurons and different activation functions. 
An example of training results for one of the variants is shown on Figure \ref{Fig_example_1}. Traffic incidents descriptions were used to predict the incident severity. The data set was split into train, validation and test sets by proportion 70:20:10. Training results show that the LSTM sentiment encoder needs at least 15 epochs to converge, so we decided to train each variant of the LSTM sentiment encoder for 15 epochs. We use Mean Squared Error (MSE) as the loss function. 




\subsubsection{The use of MSE versus cross-entropy}

MSE is a legitimate metric for the classification when the target feature is represented as an ordered variable \cite{gaudette2009evaluation} in which MSE is preferred instead of the Cross-Entropy (CE) loss in order to reduce the model complexity and the probability of over-fitting. In our research we determined that CE required Nx5 sized matrix for the intermediate feature vector to the target value classification, while the MSE solution requires only Nx1 matrix, where N is size of the intermediate feature vector). MSE loss is also superior to CE loss for class-imbalanced datasets \cite{kato2021mse} and our incident severity feature distribution poses an imbalanced classification problem.

\subsection{Artificial Neural Network Encoder for the traffic flow/speed encoding}\label{ANN_ae}

\begin{figure}[!h]
\centering
\includegraphics[width=0.4\textwidth]{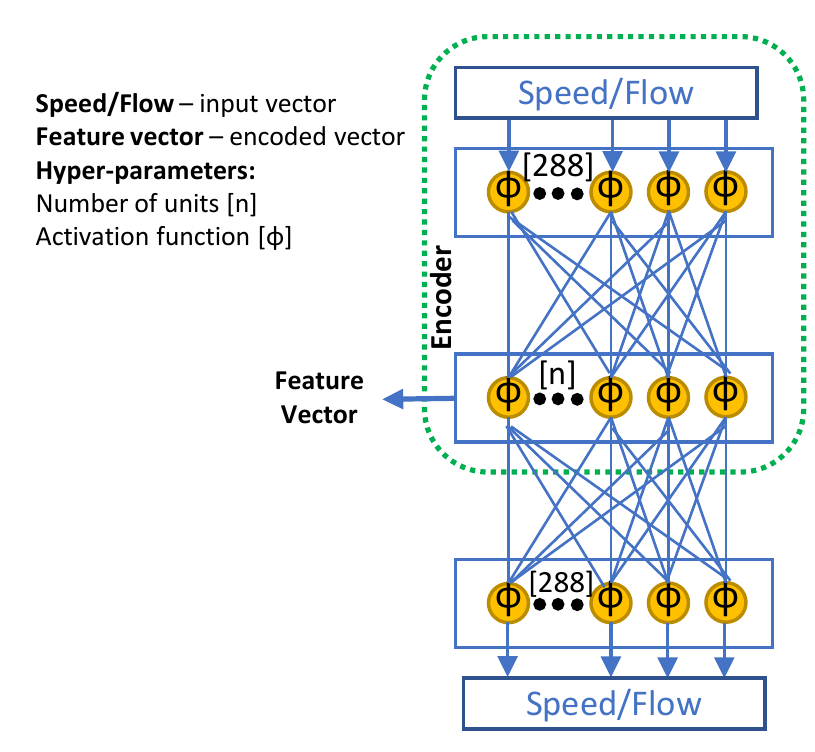}
\centering
\caption{The structure of the ANN autoencoder}
\label{Fig_ann}
\vspace{-0.3cm}
\end{figure}

As additional data sources apart from the incident baseline features, we use the general structure of Artificial Neural Network (ANNs) Autoencoder \cite{kramer1991nonlinear} with varying number of neurons and different activation functions in the bottleneck layer to produce the encoded speed/flow data sets. Flow and speed values are normalised to the corresponding maximum observed traffic speed and flow in the data set. To improve the performance of the encoding model we use all the time series data available for each incident which could be matched to a VDS station. We combine normalised flow and speed data sets to perform the ANN model training which allows the model to grasp actual time series without focusing on speed and flow on an individual level. We do make the observation that while speed and flow could be used as raw features in any ML prediction framework, the benefit of using ANN for auto-encoding is mainly a dimensionality reduction and improved accuracy in case of extreme outliers. Last, we extract the outputs of the ANN autoencoder bottleneck layer and use them as features in the ML models shown in Fig. \ref{Fig_ann}.

The following activation units were used in the bottleneck layers of the ANN autoencoder and the LSTM sentiment encoder: a) the Rectified Linear Unit (ReLU) \cite{agarap2018deep} which is a piecewise linear function (output values are $[0;+\infty]$ b) the Exponential Linear Unit (ELU) \cite{trottier2017parametric}, which was developed to reduce bias shift (which leads to weight oscillations) c) the Tanh - a hyperbolic tan function which has the property of equalizing training over layers \cite{kalman1992tanh}; its output can take values in the interval $(-1;+1)$ d) the Sigmoid activation function which output can take values in the interval $(0;1)$.

\subsection{Baseline Machine Learning model selection}

When all encoding has been finalised, we first use the following ML regression models as a baseline to perform the incident duration prediction:

a) gradient boosting decision trees - GBDT \cite{Xia2017TrafficFF} which rely on training a sequence of models, where each model is added consequently to reduce the residuals of prior models;  b) extreme gradient decision trees - XGBoost \cite{chen2015xgboost} which rely on an exhaustive search of split values by enumerating over all the possible splits on all the features and contains a regularisation parameter in the objective function; c) random forests - RF \cite{8283291} which applies a bootstrap-aggregation (bagging, which consists of training models on randomly selected subsets of data) and uses the average (or majority of votes) of multiple decision trees in order to reduce the sensitivity of a single tree model to noise in the data d) Support Vector Regression (SVR) machines \cite{drucker1997support} which are characterized by the use of kernels and symmetrical loss function (equal penalization of high and low errors), e) Decision Trees (DT) regression models \cite{breiman1984cart} which rely on the repetitive process of splitting and generates a set of rules which can be used for the value prediction, f) Linear Regression (for which we use standard Ordinary Least Squares optimisation) which represents the relation between features and the target variable as a linear equation targeting to minimize the residual sum of squares between the actual and the predicted values of the target variable.


\subsubsection{Model performance evaluation}\label{eval}
To evaluate the regression models on the task of the incident duration prediction we use the mean absolute percentage error and the root mean squared error defined as:
\begin{equation}
    MAPE = \frac{1}{n} \sum_{t=1}^{n} \left|\frac{A_t-F_t}{A_t} \right|
\end{equation}
\begin{equation}
    RMSE = \sqrt{\frac{1}{n} \sum_{i=1}^{n} (A_t-F_t)^2}
\end{equation}
where $A_t$ are the actual values and $F_t$ - the predicted values, $n$ - the number of samples. We do make the observation that other performance metrics have been obtained (MAE, SMAPE), but given the current page limitations, we focus on MAPE, RMSE results only. 

\subsubsection{Hyper-parameter tuning for the proposed regression model}\label{hyper}

We use 10-fold cross-validation to overcome the over-fitting problem \cite{geisser1993vol} and to assess the generalization performance of the ML models. In each scenario, the data set is partitioned into 10 folds. The ML regression model is trained on 9 folds to make prediction on the remaining fold. The procedure is then repeated 10 times and the accuracy results are averaged across several repetitions.

\subsection{MAPE versus RMSE comparison and their non-linear relationship}
\begin{figure*}[h!]
\centering
\includegraphics[width=0.3\linewidth]{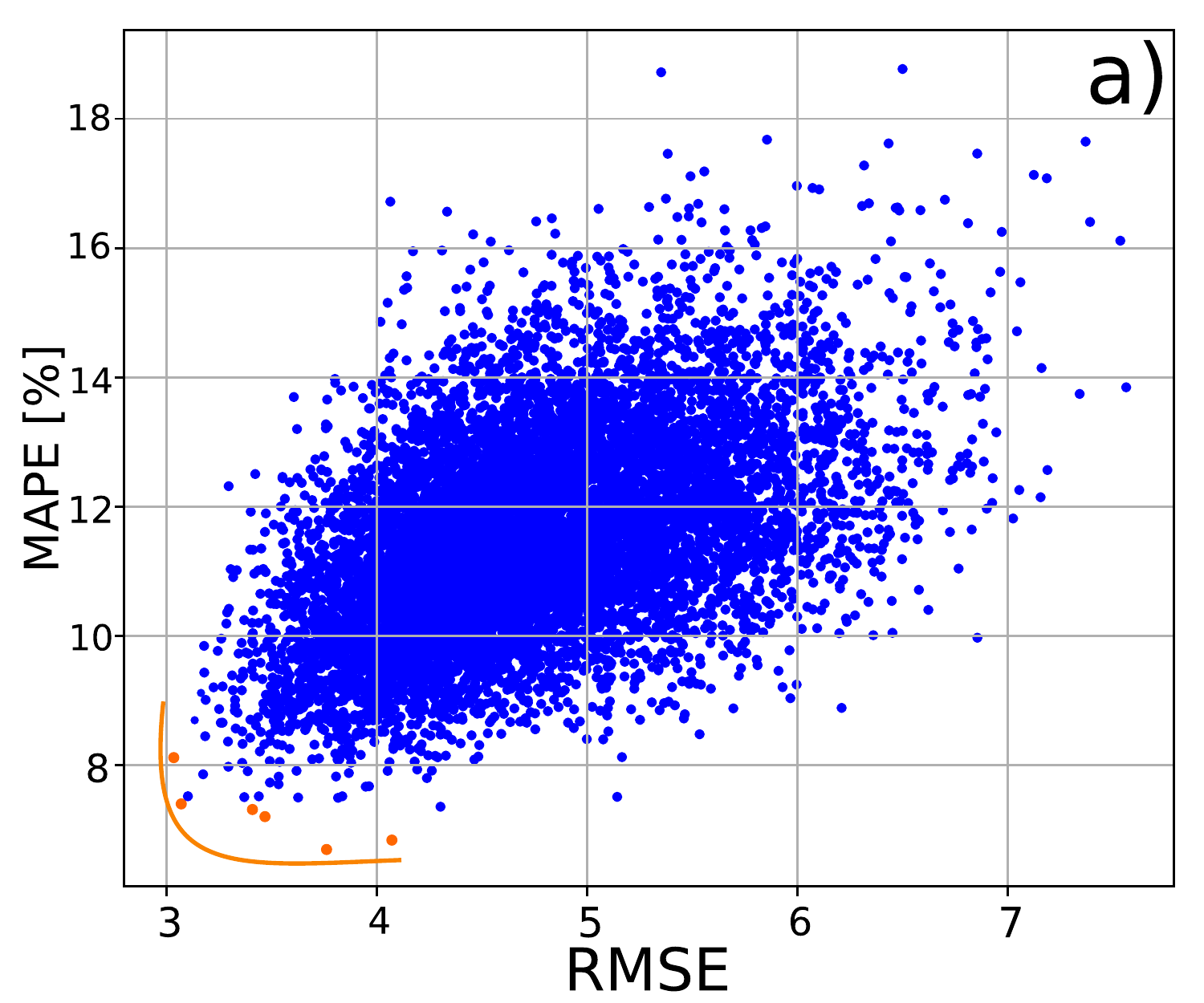}
\includegraphics[width=0.3\linewidth]{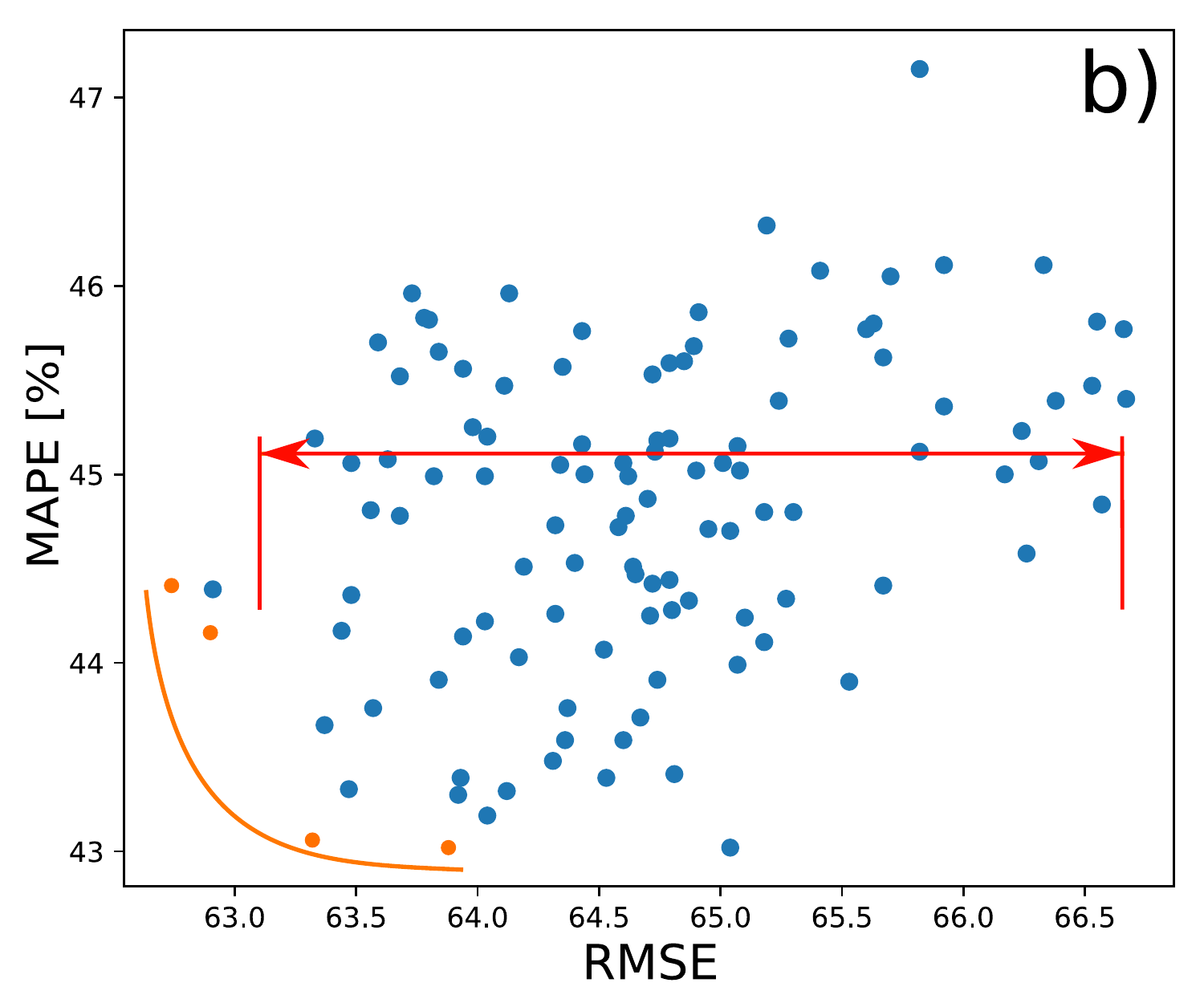}
\includegraphics[width=0.315\linewidth]{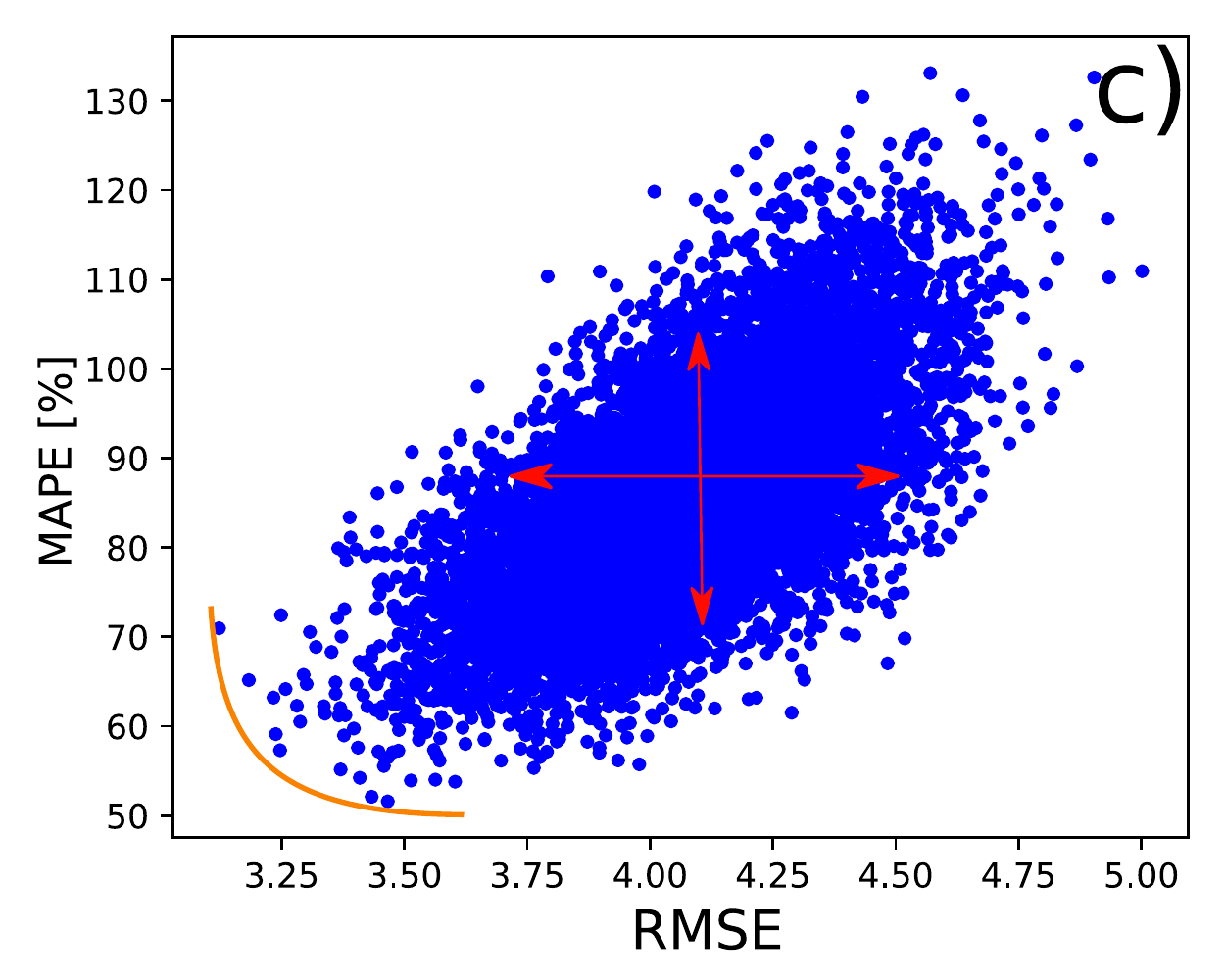}
\caption{RMSE vs MAPE results for a) CCS data set, b) CTADS - incident duration c) Random vectors}
\label{fig_A}
\vspace{-0.5cm}
\end{figure*}

There is a non-linear relationship between MAPE and RMSE when performing regression, which can be verified by using different regression data sets. We tested this hypothesis on the Concrete Compressive Strength (CCS) Data Set from UCI Machine Learning Repository by using 1000 evaluations of random 9:1 train-test splits using Random Forest evaluated against MAPE and RMSE.
\cref{fig_A}a) presents the MAPE versus RMSE plot in which we observe that, the same MAPE result (e.g. 12\%) may be attributed to multiple RMSE results (e.g. from 3.5 to 6.5). A similar situation observed for 45\% of MAPE on CTADS using Random Forest (see \cref{fig_A})b). 
Therefore, the occurrence of a higher RMSE error when MAPE becomes lower (as in our paper) and vice-versa is a correct result. MAPE vs RMSE compared between 100-units random vectors with 1-10 value interval using 10,000 evaluations (see \cref{fig_A})c). As can be seen from all three sub-plots, the decrease in MAPE doesn't necessarily mean a decrease in RMSE. 
For our study  we focused on discussing the MAPE metric, which is widely used in the literature on the topic of incident duration prediction since its intuitive meaning (e.g. a 30\% MAPE means a 30\% deviation of prediction from the actual incident duration) and a less inclination to high errors from outliers such as the case of RMSE. 
The results are part of the optimal Pareto Front [marked in orange] which showcases that our proposed method can obtain the set of optimal feature combination scenarios rather than only one winning scenario.
To conclude, despite an assumption on linear dependence between the RMSE and the MAPE metrics (assumption that both metrics should be reduced in an efficient solution), both in our incident duration case and the CCS data set, we observe a Pareto front of efficient solutions (no solution is sufficient in both metrics, making our results stand strong).

\subsection{Comparison to other baselines}
It is hard to perform a comparison between different studies on the traffic incident duration prediction since different data sets are used for research purposes \cite{li2018overview}. Majority of these data sets are also private and rely on different sets of features. CTADS data set appeared only recently (2019) and there is still no uniform convention on which data subset to use as a baseline, since the data set is big (1.5 million records) and heterogeneous (it includes reports from all kinds of traffic networks around United States). Indeed, in our previous work we have compared various ML-DL approaches against logs from Australia and USA, which can be used as extended results.

\section{RESULTS}\label{results}

\subsection{Best model selection}

\begin{figure}[ht]
\centering
\includegraphics[width=0.27\textwidth]{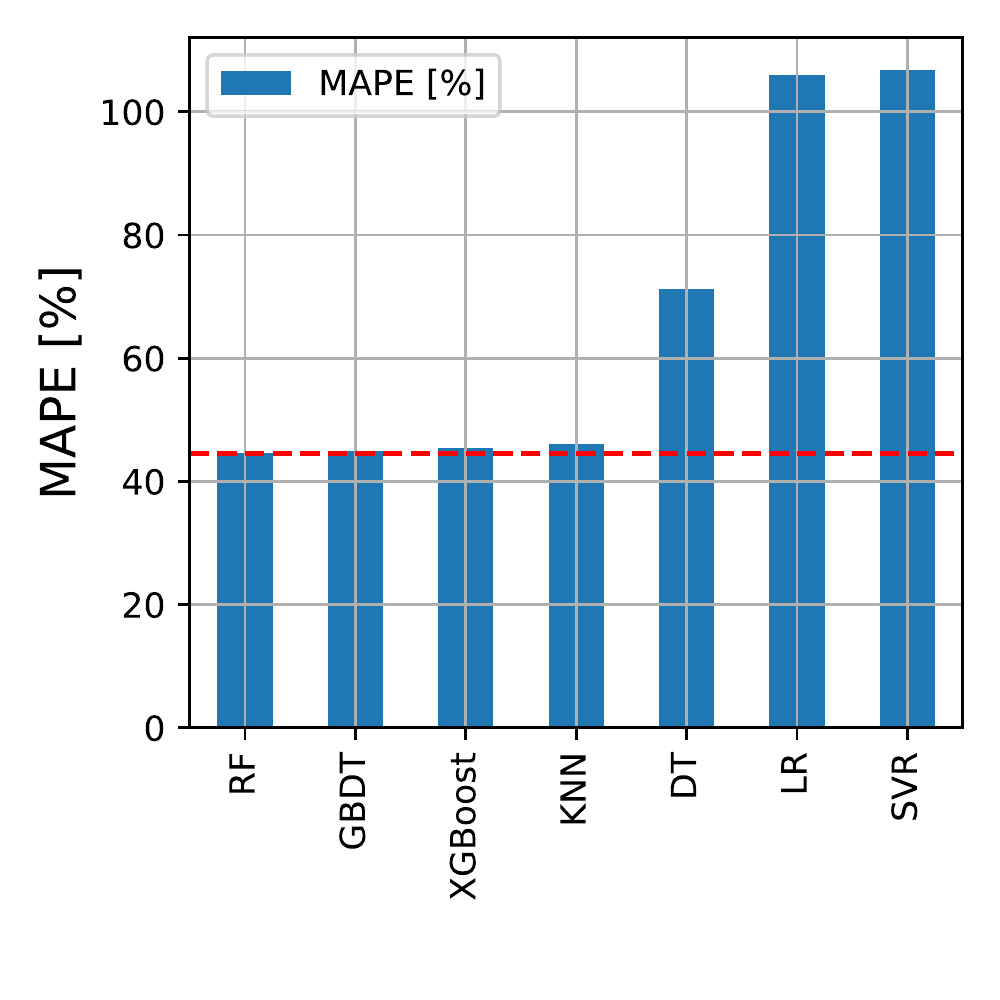}
\centering
\caption{Regression results for baseline feature set across different ML models.}
\label{figpipe}
\end{figure}

 
First, we try to find the three best models which show high performance of the baseline feature set consisting of traffic accident reports for which we have available traffic flow counter data. We do so by performing a cross-validation as described in \ref{eval} and a performance evaluation as detailed in \ref{hyper}. Figure \ref{figpipe} shows the average MAPE score for the 10-fold cross-validation obtained across several ML models such as Random Forests (RF), GBDT, XGBoost, kNN, Decision Trees (DTs), Linear Regression (LR) and Support Vector Regression (SVR). Given that the majority of traffic incident duration prediction methods published previously have reported a MAPE score below $50\%$ \cite{li2018overview}, we select RandomForest, GBDT and XGBoost as the best performing models as their MAPE score falls below $46\%$. Next, we evaluate these three models against the baseline feature set when we apply our novel modelling approach as previously explained in sections \ref{LSTM_ae}-\ref{ANN_ae}: traffic flow, speed via ANN autoencoding and textual incident description via LSTM sentiment encoding.

There are in total 140 scenarios describing combinations of additional features [7 speed/flow/text features x 5 unit count x 4 activation functions] for each of the top three ML models. Given the restricted space allocation for this article, in Tables \ref{gbdtres}-\ref{xgbres} we present only the top 8 best scenario results ranked against MAPE for each ML model. 
\begin{table}
\centering
\small
\begin{tabular}{|l|r|l|r|r|}
\toprule
AdditionData &  units & activation &  MAPE &  RMSE \\
\midrule
    baseline &      &        & 44.99 & 58.4 \\
\midrule
    LSTM-sent &     12 &       relu & 41.89 & 65.03 \\
       Flow7 &      8 &       tanh & 41.92 & 64.61 \\
    LSTM-sent &     16 &       tanh & 42.05 & 65.04 \\
      Speed7 &     16 &       tanh & 42.28 & 63.97 \\
    LSTM-sent &      8 &       tanh & 42.43 & 64.13 \\
    LSTM-sent &     16 &        relu & 42.56 & 65.82 \\
        Flow &     16 &    sigmoid & 42.57 & 64.53 \\
      Speed7 &      2 &    sigmoid & 42.59 & 64.76 \\
\bottomrule
\end{tabular}
\caption{Top 8 best scenario results for GBDT-enabled framework}\label{gbdtres}
\vspace{-0.5cm}
\end{table}

Findings reveal that the encoded textual description is among the top 3 configurations for every regression model as seen from Tables \ref{gbdtres}-\ref{xgbres}. Models also demonstrate a preference for the way of encoding: 1) the Tanh activation function forms a majority in the top results for GBDT both for encoding the incident description and flow/speed features (Table \ref{gbdtres}), 2) the ReLU activation function forms a majority in the case of XGBoost (Table \ref{xgbres}). This observation can point on a preference in the way of encoding features when using specific regression models. The best performing model among the top three finalists, when using all additional features seems to be GBDT: the best results are obtained when encoding the traffic incident description and when using the traffic flow 7 days before the incident with 12 units and the ReLU activation function [$MAPE=41.89\%$, Table \ref{gbdtres}] (therefore including the information on the regular traffic flow profile on the same weekday, together with the incident report proves important for the task of incident duration prediction). 

\begin{table}
\centering
\small
\begin{tabular}{|l|r|l|r|r|}
\toprule
AdditionData &  units & activation &  MAPE &  RMSE \\
\midrule
        baseline &       &        & 44.58 & 57.6 \\
\midrule
        Flow &      4 &       tanh & 43.02 & 63.88 \\
    LSTM-sent &      4 &        elu & 43.02 & 65.04 \\
    LSTM-sent &     12 &       relu & 43.06 & 63.32 \\
       Flow7 &     16 &    sigmoid & 43.19 & 64.04 \\
        Flow &     16 &        elu & 43.30 & 63.92 \\
          FD &      4 &        elu & 43.32 & 64.12 \\
       Flow7 &     16 &       tanh & 43.33 & 63.47 \\
          FD &      4 &    sigmoid & 43.39 & 64.53 \\
\bottomrule
\end{tabular}
\caption{Top 8 best results for RF}\label{rfres}
\vspace{-0.5cm}
\end{table}

Other models show a higher MAPE or RMSE results for the incident duration prediction (see RF enabled results in Table \ref{rfres} with lowest $MAPE=43.2\%$ for a combination of baseline, regular traffic flow, 4 layer units and a tanh activation function); similar findings appear for XGBoost-enabled results in Table \ref{xgbres} with the lowest $MAPE=43.44\%$, when using again the regular flow features, 8 layer units and ReLU activation function. This experiment shows that an accurate incident duration prediction immediately after the event has occurred is possible, leveraging the incident description and the measured traffic flow on the day of accident, which may prove very useful for TMCs to incorporate directly in their incident management platforms. Lower MAPE does not necessarily mean lower RMSE as seen from the baseline and additional data scenarios, but the LSTM sentiment encoding seems to be the approach that obtains the best RMSE score (64.13) when combined indeed with other variations of the activation function and number of hidden units (as shown in Table \ref{gbdtres}). 
\begin{table}
\centering
\small
\begin{tabular}{|l|r|l|r|r|}
\toprule
AdditionData &  units & activation &  MAPE &  RMSE \\
\midrule
        baseline &       &        & 45.44 & 63.41 \\
\midrule
        Flow &      8 &       relu & 43.44 & 69.93 \\
    LSTM-sent &      4 &       tanh & 43.58 & 71.03 \\
      Speed7 &     16 &       tanh & 43.63 & 71.62 \\
          SD &      4 &       relu & 43.73 & 70.58 \\
      Speed7 &     16 &       relu & 43.80 & 71.92 \\
    LSTM-sent &     16 &        elu & 43.81 & 70.45 \\
    LSTM-sent &      8 &       relu & 43.82 & 72.19 \\
       Flow7 &      2 &       relu & 43.85 & 72.94 \\
\bottomrule
\end{tabular}
\caption{Top 8 best results for XGBoost}\label{xgbres}
\vspace{-0.5cm}
\end{table}

\subsection{Parallel coordinates for scenario setup}

\begin{figure*}[!ht]
\centering
\includegraphics[width=0.6\textwidth]{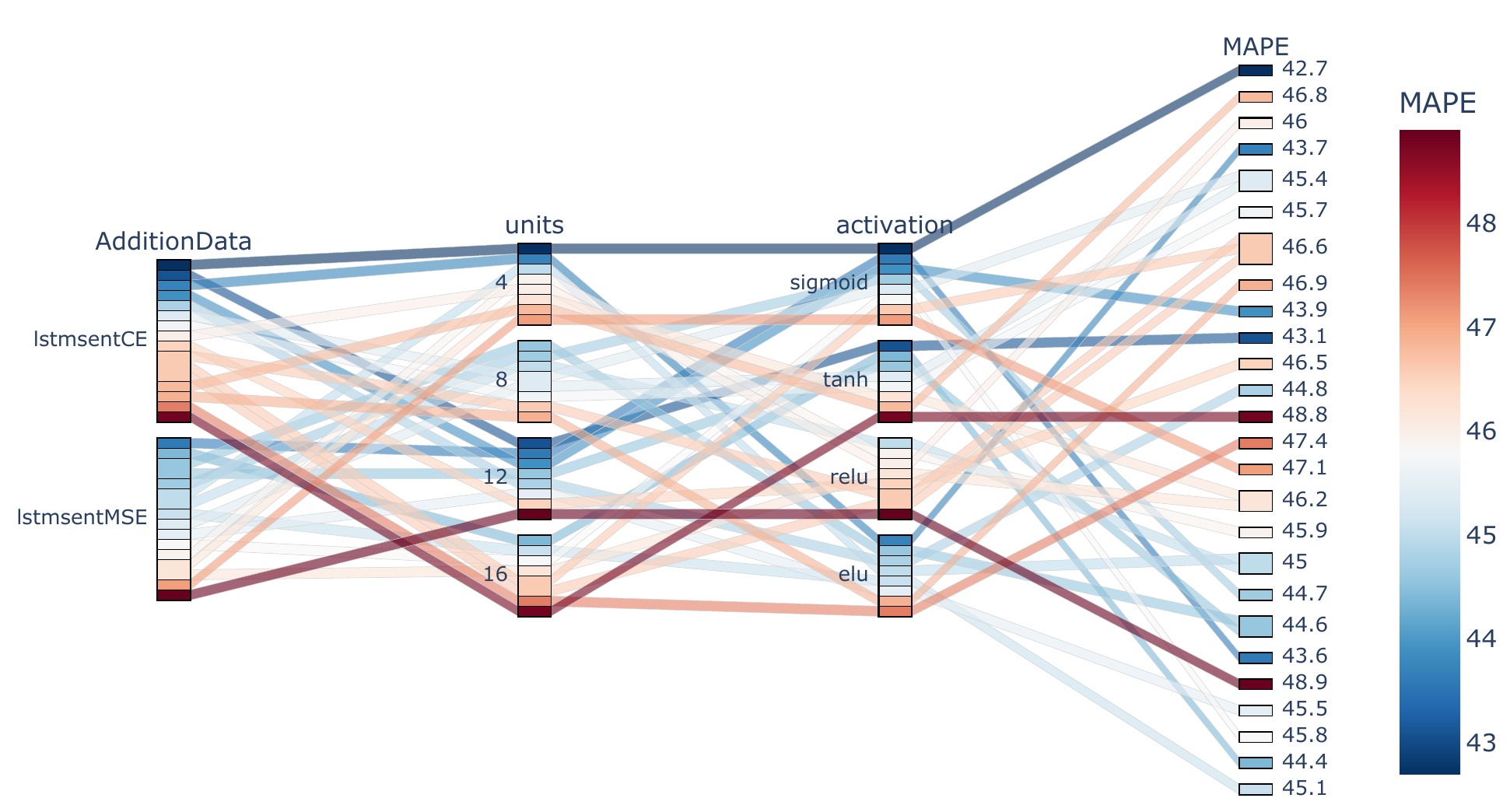}
\centering
\caption{Parallel categories representation for all regression scenarios with GDBT.}
\label{GBDT_parallel_cats}
\vspace{-0.5cm}
\end{figure*}

To supplement the findings, we also provide a parallel categories representation of all the 140 scenarios for the GBDT model in Figure \ref{GBDT_parallel_cats}, which highlights the best combination of activation functions that seem to be working best alongside the character-level LSTM sentiment encoder of traffic flow incident textual description and speed information - mostly from previous daily speed profiling using historical data. The worst results seem to be the ones obtained when using only the speed or flow difference vector alongside the baseline incident features.

Encoding using Sigmoid and Tanh activation units on average performs best, probably because of the limited value range: Tanh an Sigmoid allow encoded representations to take values in ranges $[-1;+1]$ and $(0;1)$ correspondingly, ReLU and ELU can take unlimited positive values. These results indicate which value ranges work best for encoded representation.


Comparison of MSE and CE implementations of lstm severity classification metric for the purpose of obtaining feature vector representation of Incident Description (see \cref{GBDT_parallel_cats}) shows that a sentiment classifier with Cross-entropy (lstmsentCE) as a target metric with one-hot encoded severity values is more efficient (left column attributed to lstmsentCE shows more blue rows associated with low metric values than lstmsentMSE - sentiment encoder which predicts severity as a single value). Comparison between the number of units shows preference for 4 units since the presence of the lowest error and absence of the highest error rows. Among the activation units, the Sigmoid is the best performer showing more low error results than other units. This scenario is to show how feature vector representing incident description may may be efficiently encoded to be used with conventional GBDT machine learning method: using cross-entropy for the severity classification, using 4 units and the Sigmoid as activation function.

\section{Conclusion}\label{conclusions}
In this paper, we have proposed a novel framework to predict the incident duration using an integration of machine learning with traffic flow and description features encoded via several Deep Learning methods. This approach demonstrates the stable and noticeable improvement across all the performing models. The results give evidence to the importance of using specific deep-learning encoding approaches for all regression models which provide a further boost-up in the model performance from past historical traffic information and the textual incident description. Efficiently encoding incident-related features for the task of incident duration prediction is the first step to model the traffic incident impact on the traffic flow. Further work is currently being focused on exploring the spatial and the temporal dynamic prediction of the incident impact via graph-based modelling approaches.
The research has the following limitations: a) we used as study area only San Francisco, but there is a data availability on traffic accidents and traffic flow for the area of California, b) traffic speed and flow were taken into account only before the incident; by collecting traffic count data for longer periods it possible to build traffic speed/flow profiles which may provide more accurate predictions.
The societal impact of the research is as follows: the data availability of the predicted incident duration can improve for TMC incident and traffic management (e.g. TMC can announce when an incident is expected to dissipate, how many resources to allocate, etc), which in turn will reduce the time spent by people in the traffic congestion caused by the incident.
The code for the paper can be found: \url{https://github.com/Future-Mobility-Lab/TIDP_2022}.






%
\section*{Acknowledgments}

This research is supported by the University of Technology Sydney (ARC LP project LP180100114), funded by iMOVE CRC and supported by the Cooperative Research Centres program, an Australian Government initiative.

%



{\footnotesize
\bibliographystyle{IEEEtran}
\bibliography{IEEE_ITSC_2022}}

\section*{Appendix}

\subsection{Word importance for severity classification}

\begin{figure}[!ht]
\centering
\includegraphics[width=0.5\textwidth]{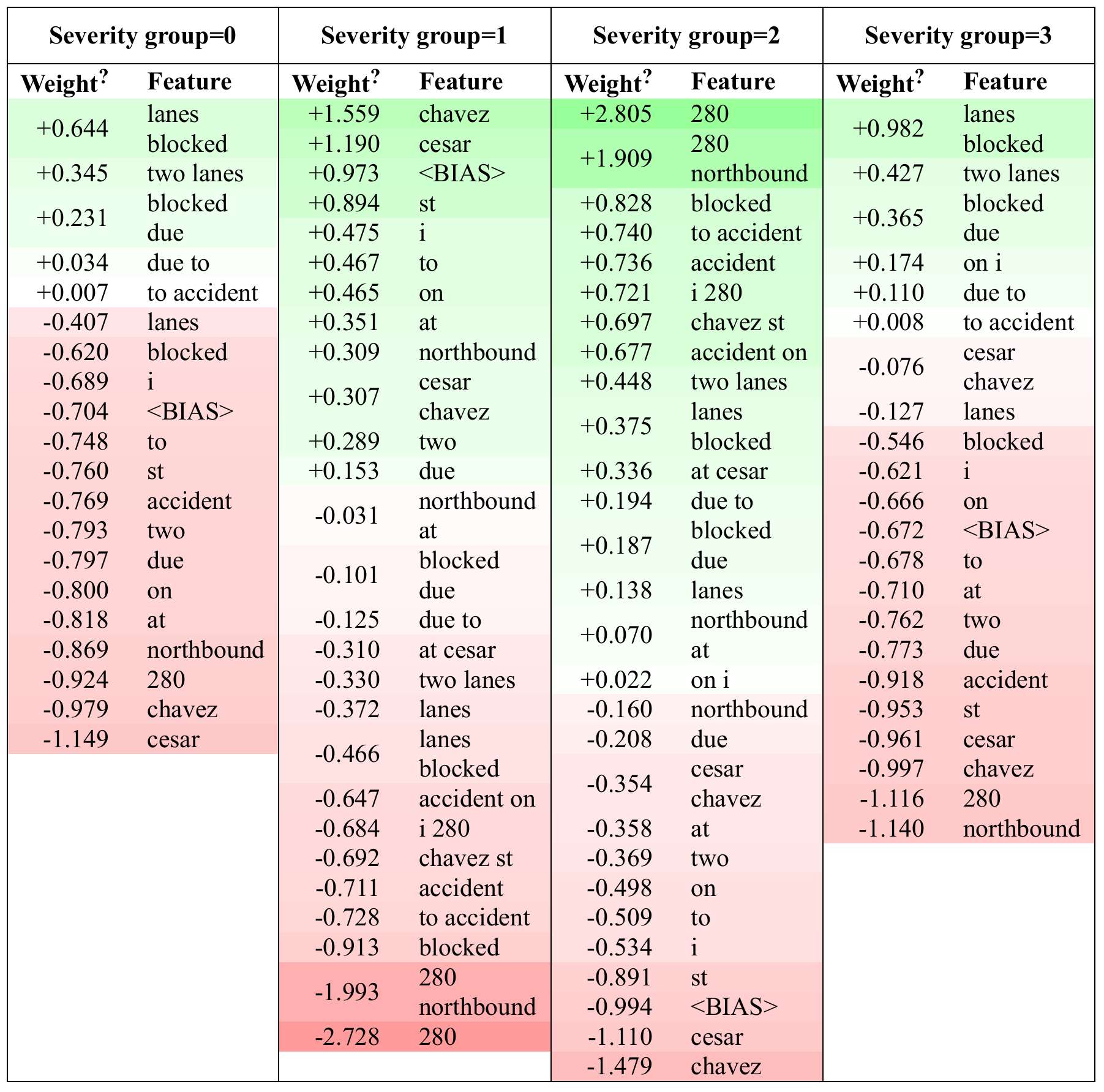}
\centering
\caption{Word importance estimation using LIME method for incident severity groups}
\label{wordimpsev}
\vspace{-1.5cm}
\end{figure}

To estimate word importance in the Incident Description feature, word count matrix has been transformed to a normalized TF-IDF representation (term frequency–inverse document frequency) \cite{TFIDF}. N-gram value range is (1,2). Then linear dimensionality reduction has been performed using truncated singular value decomposition to 50 componenets for 7 iterations. Then we used GBDT classification model to fit incident severity and three quantiled groups (ratio 33\%:33\%:33\% to represent equaly sized groups with duration intervals 0-29min, 30-71min and 72-2750min) of the incident duration. Classifer predictions were then analyzed for feature importance using LIME method \cite{LIME}, where every feature represents 1 word or 2 word combination presence in the incident description.
One or more combinations of word in the description can contribute to the incident being classifed into one of severity groups (\cref{wordimpsev}) - presence of "lanes blocked" and "two lanes blocked" has the highest contribution to the incident being classifed into highest (3) or lowest (0) severity group. Severity 1 or 2 is more related to the actual location, which represented as word describing Cesar Chavez St and I-280 Interstate Highway. High positive and opposite high negative contribution of words towards severity group observed for severity groups 1 and 2, where "280" and "chavez" have high opposite contributions, making this groups easily separable. 
When we perform classification towards equaly sized incident duration groups, "lanes blocked" has the highest positive contribution of the incident to be classified into low duration group. If accident happens on Cesar Chavez St, it can be easily classified into low duration group signifying importance of location for the task of incident duration prediction. High negative contribution of "lanes blocked" observed for duration group 1 with the highest contribution of "280" word meaning that incident appears on I-280 Interstate Highway.

\begin{figure}[!ht]
\centering
\includegraphics[width=0.5\textwidth]{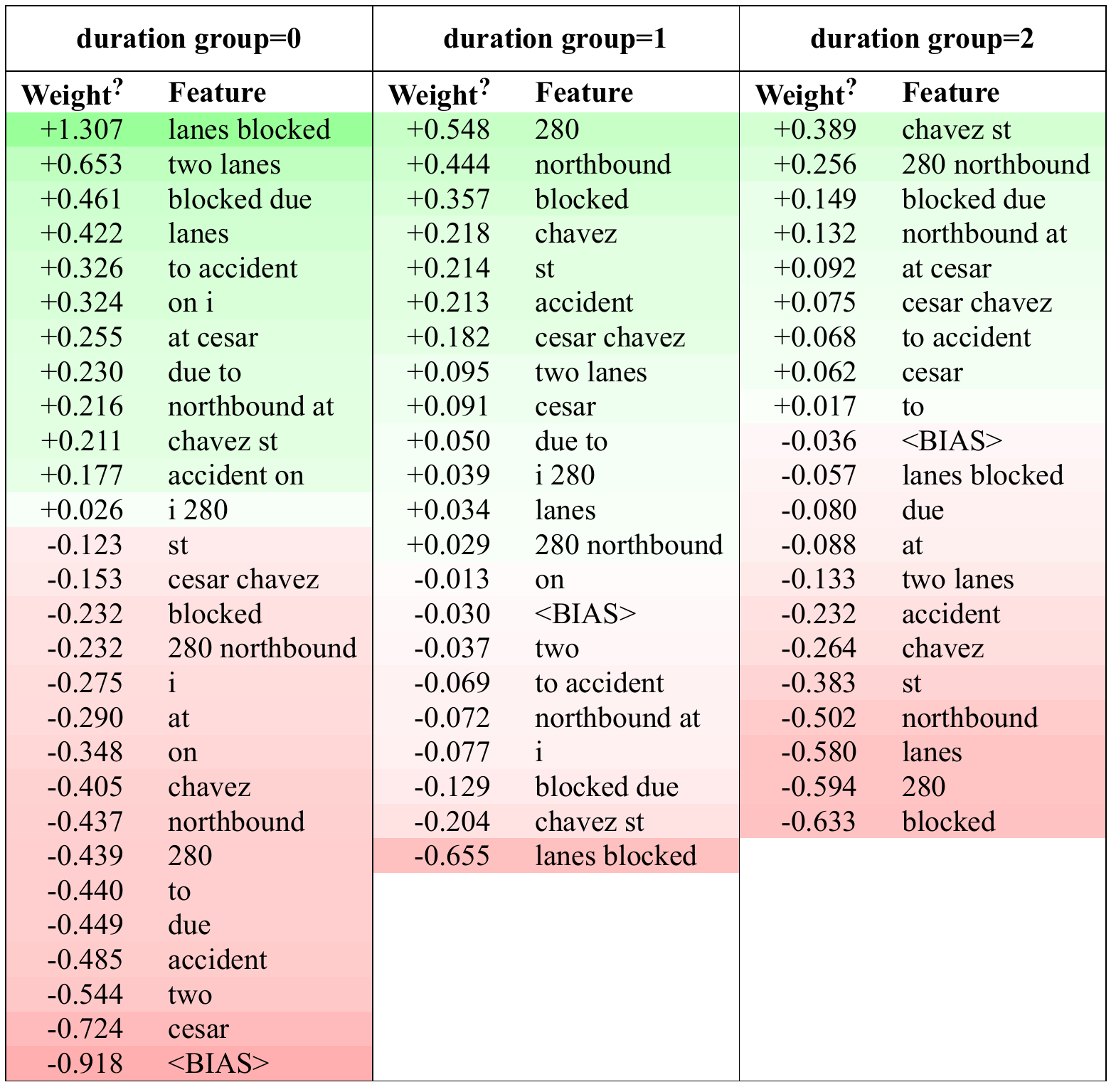}
\centering
\caption{Word importance estimation using LIME method for incident duration groups}
\label{wordimpdur}
\end{figure}

\subsection{Traffic flow and traffic speed on the day of the incident}

\begin{figure*}[!ht]
\centering
\includegraphics[width=\textwidth]{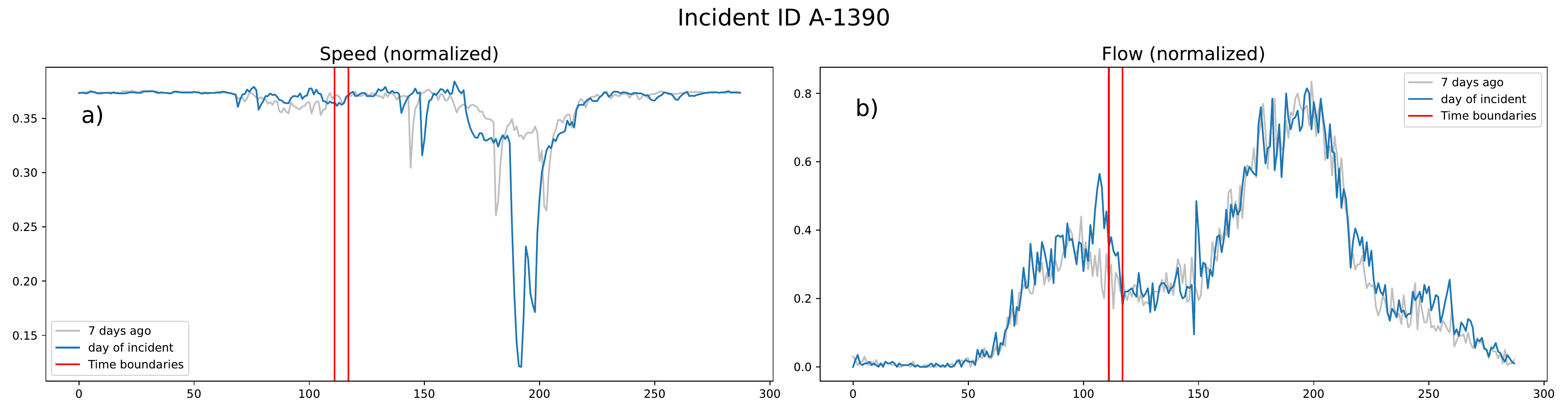}\\
\includegraphics[width=\textwidth]{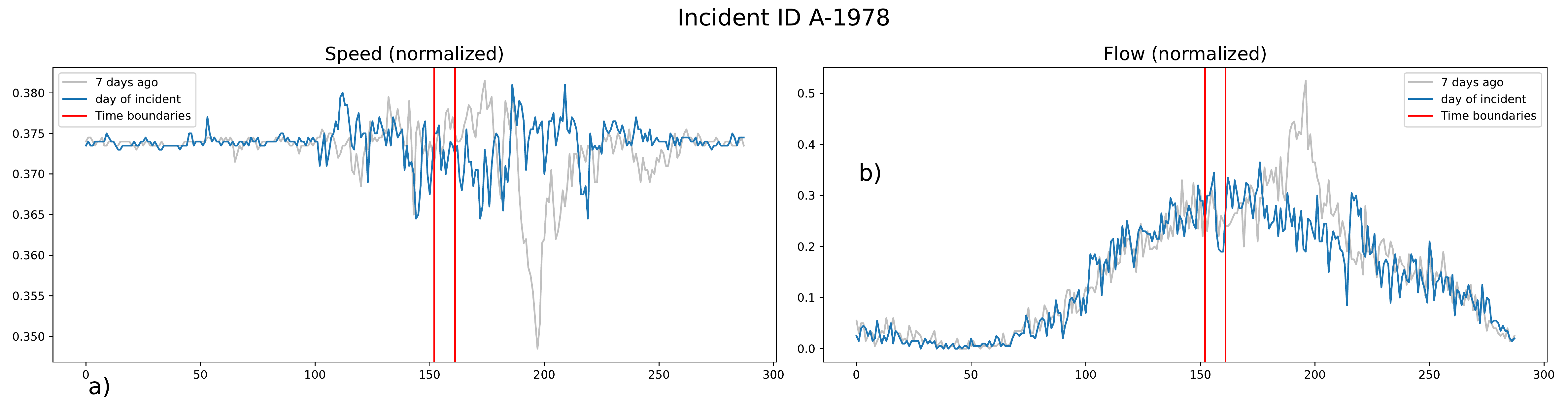}\\
\includegraphics[width=\textwidth]{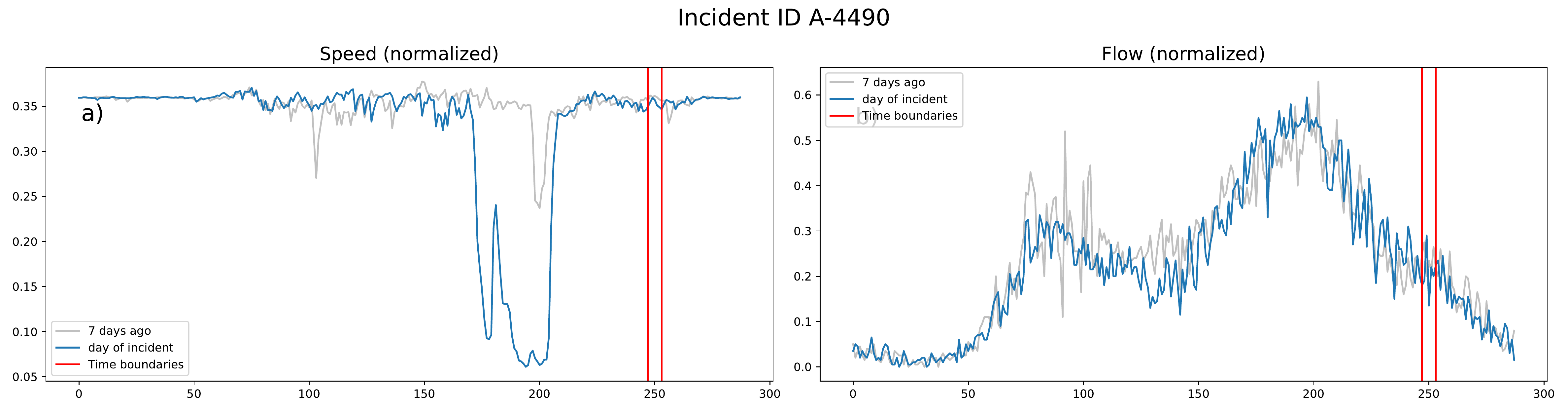}\\
\includegraphics[width=\textwidth]{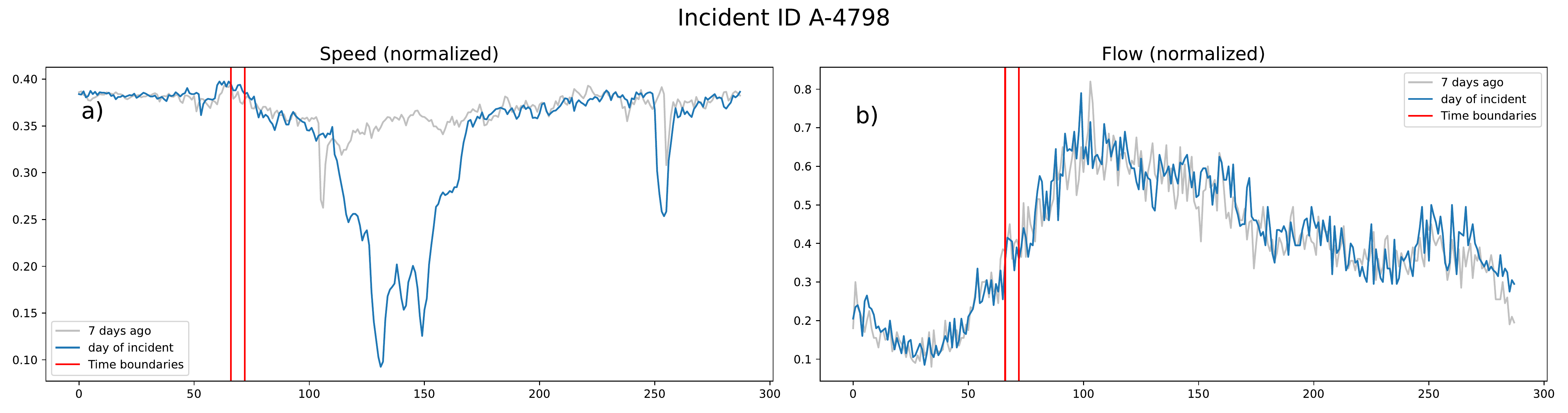}\\
\includegraphics[width=\textwidth]{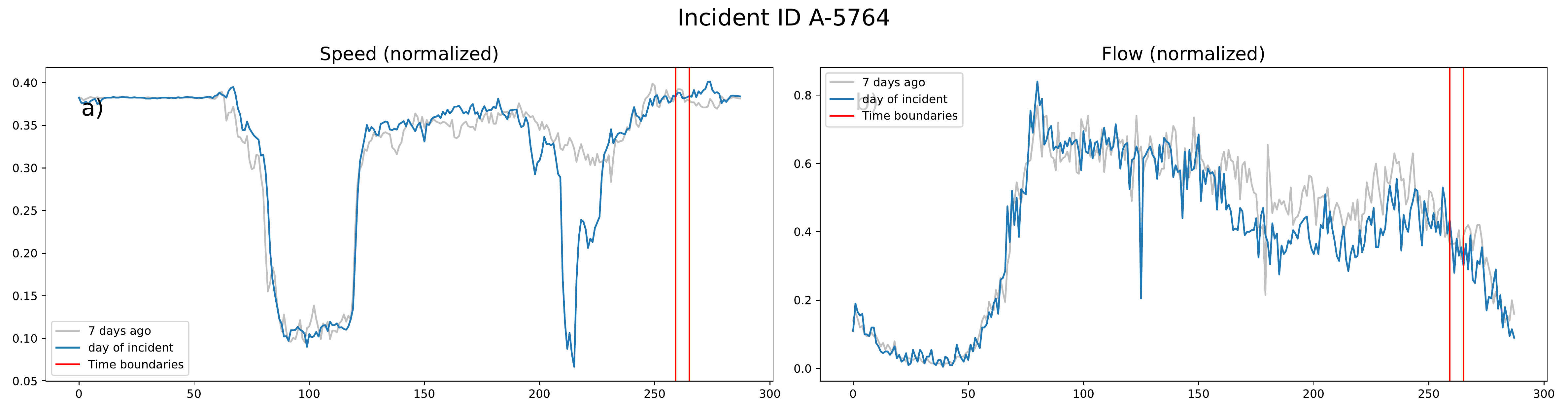}\\
\centering
\caption{Traffic speed and flow during the day of the incident. Part \#1}
\label{wordimpdur1}
\end{figure*}

\begin{figure*}[!ht]
\centering
\includegraphics[width=\textwidth]{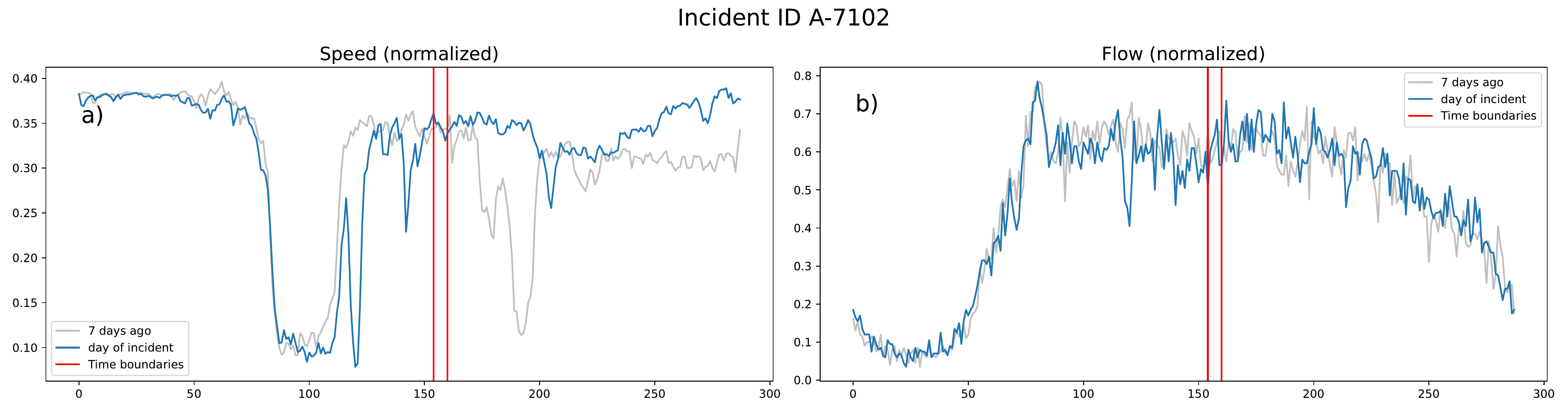}\\
\includegraphics[width=\textwidth]{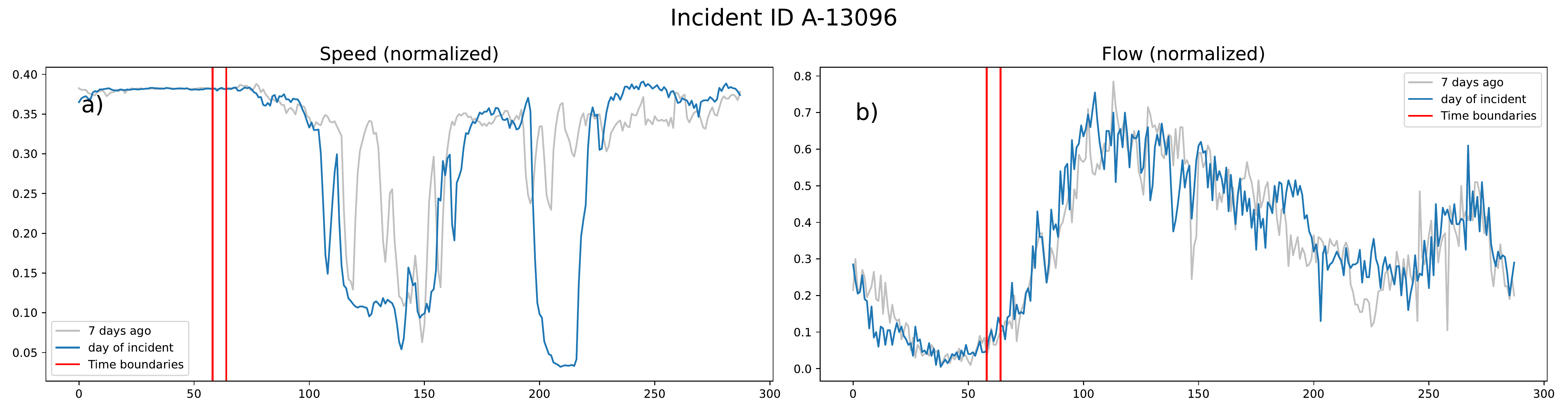}\\
\includegraphics[width=\textwidth]{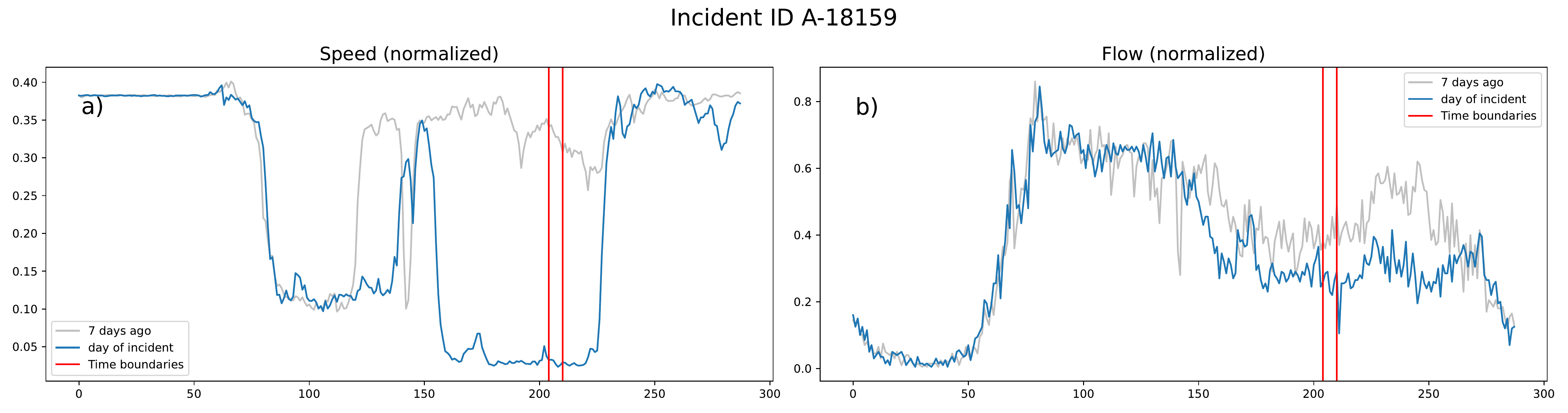}\\
\includegraphics[width=\textwidth]{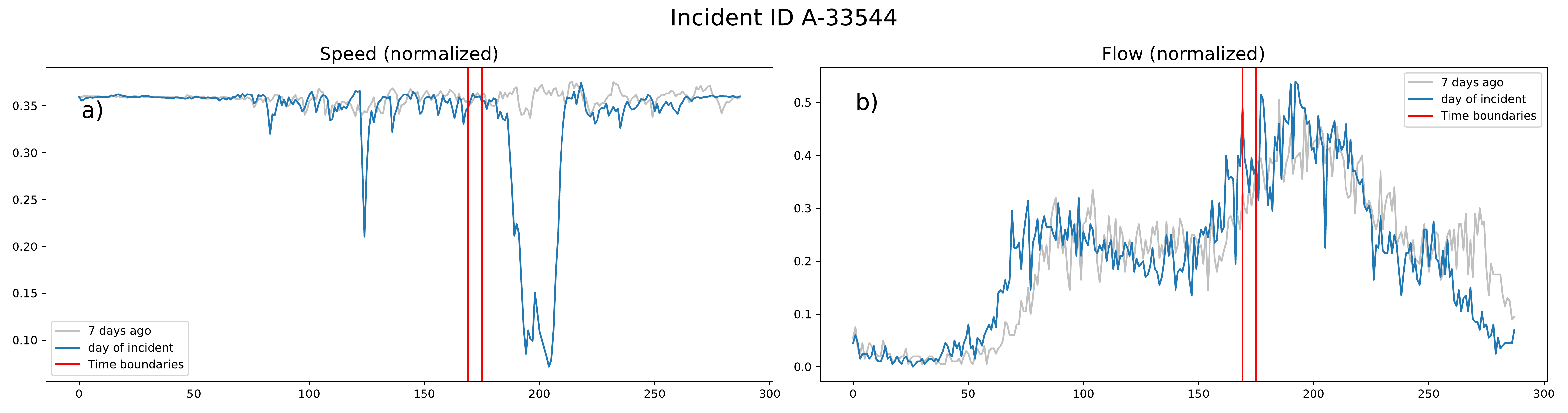}\\
\includegraphics[width=\textwidth]{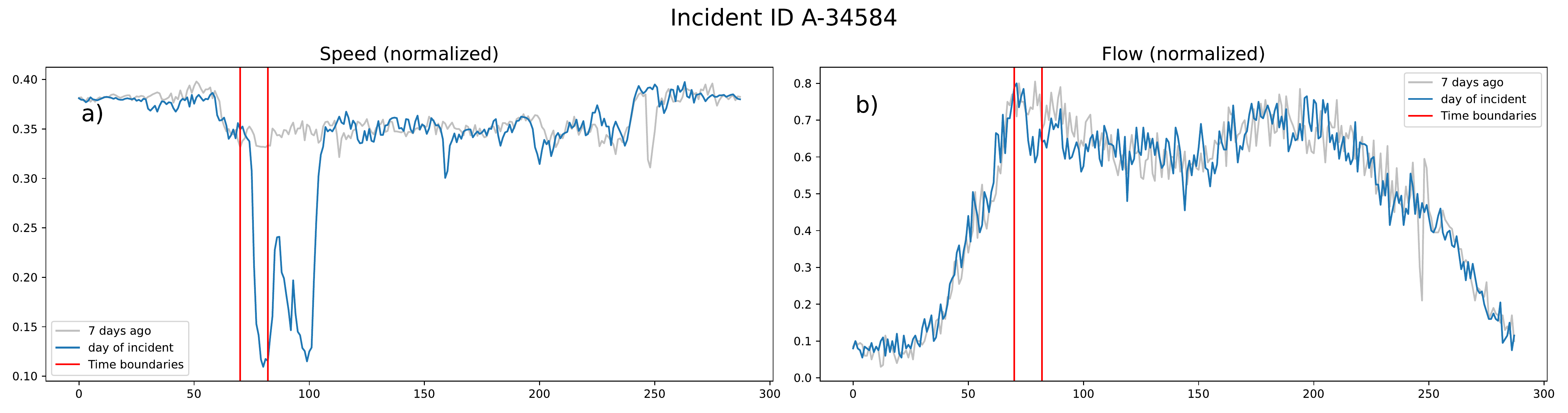}\\
\centering
\caption{Traffic speed and flow during the day of the incident. Part \#2}
\label{wordimpdur2}
\end{figure*}

The following plots represent recorded traffic speed and flow on the day of the incident and week before in 500m proximity of the incident along the road (see Fig. \ref{wordimpdur1} and \ref{wordimpdur2}). Reports in CTADS data set indicate that the highest impact of traffic incident is attributed to significant decrease in traffic speed, while traffic flow stays the least affected by disruption.

%
%
%

\end{document}